\documentclass[12pt,a4paper]{article} 
\usepackage{amsmath,amssymb}
\usepackage{mathptmx} 
\usepackage{amsfonts,xcolor}
\usepackage{graphicx}
\usepackage{caption}
\usepackage{subfigure}
\usepackage{algorithm2e}
\usepackage{abstract}

\newcommand{\Rr}{{\mathbb R}}

\title{A Novel Active Contour Model for Texture Segmentation}

\author{Aditya Tatu\footnote{aditya.tatu@gmail.com} \and
        Sumukh Bansal\footnote{sumukh\_bansal@daiict.ac.in} }

\begin{document}


\maketitle
\begin{abstract}
Texture is intuitively defined as a repeated arrangement of a basic pattern or object in an image. There is no mathematical definition of a texture though. The human visual system is able to identify and segment different textures in a given image. Automating this task for a computer is far from trivial.\\
There are three major components of any texture segmentation algorithm: (a) The features used to represent a texture, (b) the metric induced on this representation space and (c) the clustering algorithm that runs over these features in order to segment a given image into different textures.\\
In this paper, we propose an active contour based novel unsupervised algorithm for texture segmentation. We use intensity covariance matrices of regions as the defining feature of textures and find regions that have the most inter-region dissimilar covariance matrices using active contours. Since covariance matrices are symmetric positive definite, we use geodesic distance defined on the manifold of symmetric positive definite matrices $PD(n)$ as a measure of dissimlarity between such matrices. We demonstrate performance of our algorithm on both artificial and real texture images.
\end{abstract}
\section{Introduction}
Texture is intuitively defined as a repeated arrangement of a basic pattern or object in an image. There is no universal mathematical definition of a texture though. The human visual system is able to identify and segment different textures in a given image without much effort. Automating this task for a computer, though, is far from trivial.\\
Apart from being a tough academic problem, texture segmentation has several applications. Texture segmentation has been applied to detect landscape changes from aerial photographs in remote sensing and GIS \cite{yu2003}, content based image retrieval \cite{fauzi2006} and diagnosing ultrasound images \cite{muzzolini1993} and others.\\
There are three major components in any texture segmentation algorithm: (a) The model or features that define or characterize a texture, (b) the metric defined on this representation space, and (c) the clustering algorithm that runs over these features in order to segment a given image into different textures.\\
There are two approaches of modeling a texture: Structural and Statistical. The structural approach describes a texture as a specific spatial arrangement of a primitive element. Vor-onoi  polynomials are used to specify the spatial arrangement of these primitive elements \cite{voronoi_polynomial,ahuja_texel}. The statistical approach describes a texture using features that encode the regularity in arrangement of gray-levels in an image. Examples of features used are responses to Gabor filters \cite{dennis_gabor}, graylevel co-occurrence matrices \cite{zucker1980,haralick1979}, Wavelet coefficients \cite{do2002}, human visual perception based Tamura features \cite{tamura1978}, Laws energy measures \cite{laws1980}, Local binary patterns\cite{ojala1996} and Covariance matrices of features \cite{tuzel_region_covariance,michael_covariance}. In \cite{howarth04}, the authors compare performance of some of the above mentioned features for the specific goal of image retrieval. In fact, Zhu, Wu \& Mumford \cite{zhu98} propose a mechanism of choosing an optimal set of features for texture modeling from a given general filter bank. Markov random fields\cite{george_mrf}, Fractal dimensions\cite{chaudhuri_fractal} and the space of oscillating functions \cite{vese2003} have also been used to model textures.\\
Various metrics have been used to quantify dissimilarity of features: Euclidean, Chi-squared, Kullback-Leibler \& its symmetrized version \cite{wang_dti}, manifold distance on the Gabor feature space \cite{sagiv2006} and others. $k-$NN, Bayesian inference, $c-$ means, alongwith active contours algorithms are some of the methods used for clustering/segmentating texture areas in the image with similar features.\\
In this paper, we use intensity convariance matrices over a region as the texture feature. Since these are symmetric positive definite matrices which form a manifold, denoted by $PD(n)$, it is natural to use the intrinsic manifold distance as a measure of feature dissimilarity. Using a novel active contours method, we propose to find the background/foreground texture regions in a given image by maximizing the geodesic distance between the interior and exterior covariance matrices. This is the main contribution of our paper. \\
In the next subsection we list out some existing texture segmentation approaches using active contours model.
\subsection{Related work}
{
Sagiv, Sochen \& Zeevi \cite{sagiv2006} generalize both, geodesic active contours and Chan \& Vese active contours, to work on a Gabor feature space. The Gabor feature space is a parametric $2-$D manifold embedded in $\Rr^7$ whose natural metric is used to define an edge detector function for geodesic active contours, and to define the intra-region variance in case of the Chan \& Vese active contours. In \cite{savelonas2006}, the authors use Chan \& Vese active contours on Local Binary Pattern features for texture segmentation.\\
In \cite{rousson2003}, the authors propose a Chand \& Vese active contour model on probability distribution of the structure tensor of the image as a feature.
The closest approach to our algorithm is by Houhou et.\;al.\cite{nawal_acm}, where the authors find a contour that maximizes the KL-divergence based metric on probability distribution of a feature for points lying inside the contour and outside the contour. The feature used is based on principal curvatures of the intensity image considered as a $2-$D manifold embedded in $\Rr^3$. In particular, the cost function for a curve $\Omega$ is defined as 
\begin{align*}
KL(p_{in}(\Omega),p_{out}(\Omega)) & =  \int_{\Rr^+} \left(p_{in}(\kappa_t,\Omega) - p_{out}(\kappa_t,\Omega)\right)\\ 
 & \cdot \left(\log p_{in}(\kappa_t,\Omega) - \log p_{out}(\kappa_t,\Omega)\right)\ d\kappa_t
\end{align*}
where $p_{in}(\Omega),p_{out}(\Omega)$ is the probability distribution of the feature $\kappa$ inside and outside the closed contour $\Omega$ respectively. Gaussian distribution is assumed as the model for the probability distribution of the feature both inside as well as outside the contour.
In our approach, instead of using some scalar feature to represent texture, we iteratively compute a contour that maximizes the geodesic distance between the interior and exterior intensity covariance matrix of the contour. It can be seen that the maximization process has to be carried out over the manifold of symmetric positive definite matrices, making it fundamentally different from the approach in \cite{nawal_acm}. Moreover, we can easily extend this approach to covariance matrices of any other texture feature one may want to use.\\
The paper is organized as follows:In next section we provide a brief review of  active contour models for image segmentation. 
In section refer{section label}, we describe our active contour model based on geodesic distance between the interior and exterior covariance matrices of a contour. We give our experimental results in Section \ref{sec:expts} followed by conclusions and future scope. 
\subsection{Active contours and Level sets}
In classical active contours \cite{kass}, user initializes a curve $C(q): [0,1]\rightarrow \Omega \subseteq\mathbb{R}^{2}$ on an intensity image $I:\Omega \rightarrow \Rr^2$ which evolves and stabilizes on the object boundary. The curve evolution is gradient descent of an energy functional, $E(C)$, given by
\begin{align*}
E(C)=\alpha\int_{0}^{1}\mid C^{'}(q)\mid^{2} dq & + \beta \int_{0}^{1}\mid C^{''}(q)\mid^{2} dq \\
& - \lambda \int_{0}^{1}\mid\nabla I(C(q))\mid dq
\end{align*}
where $\alpha,\beta$ and $\lambda$ are real positive constants, $C^{'}$ and $C^{''}$ are first and second derivatives of $C$ and $\nabla I$ is the image gradient.  First two terms are regularizers, while the third term pushes the curve towards the object boundary.\\
Geodesic active contours \cite{kimmel} is an active contour model where the objective function can be interpreted as the length of a curve $C:[0,1] \rightarrow \Rr^2$ in a Riemannian space with metric induced by image intensity. The energy functional for geodesic active contour is given by 
\[
E=\int_{0}^{1} g(\vert \nabla I(C(q)\vert)\vert C'(q) \vert dq,
\]
where $g:\Rr\rightarrow\Rr$ is a monotonically decreasing \emph{edge detector} function. One such choice is $g(s) = \exp\left(-s\right)$.
The curve evolution equation that minimizes this energy is given by 
\[
\frac{\partial C}{\partial t} = \left(g(I)\kappa - \langle\nabla g, \hat{n}\rangle\right)\hat{n}
\]
where $\hat{n}$ is the inward unit normal and $\kappa$ is the curvature of the curve $C$.\\
A convenient computational procedure for curve evolution is the level set formulation \cite{problems_in_image_processing}. Here the curve is embedded in the zero set of a function $\phi:\mathbb{R}^{2}\rightarrow\mathbb{R}$ and the function is made to evolve so that its zero level set evolves according to the desired curve evolution equation. 
For a curve evolution equation of the form $\frac{\partial C}{\partial t}=v \hat{n}$, the corresponding level set evolution is $\frac{\partial \phi}{\partial t}=v \vert\nabla \phi \vert$. See Appendix in \cite{kimmel}. In particular the level set evolution for geodesic active contours is given by
\begin{align*}
\frac{\partial \phi}{\partial t}=g(I)\vert\nabla\phi\vert div\left(\frac{\nabla\phi}{\vert\nabla\phi\vert}\right)+ \nabla g(I).\nabla\phi.
\end{align*}
Another active contours approach was introduced by Chan and Vese \cite{chan2001} where the energy function was based on regional similarity properties of an object, rather then its edges (image gradient). Suppose that $C$ is the initial curve defined on the domain $\Omega$ of the intensity image $I$. $\Omega$ can be divided into two parts, interior (denoted by $int(C)$) and exterior (denoted by $ext(C)$). 
Let us represent the mean gray value of the region $int(C)$ and $ext(C)$ by $c_{1}$ and $c_{2}$ respectively, then the energy function for which the object boundary is a minima is given by 
\begin{align*}
F_{1}(C)+F_{2}(C) & = \int_{int(C)}\vert I(x,y)-c_{1}\vert^{2} dx dy\\
& + \int_{ext(C)}\vert I(x,y)-c_{2}\vert^{2} dx dy 
\end{align*}
After adding some regularizing terms the energy functional $F(c_{1},c_{2},C)$, is given by
\begin{align*}
F(c_{1},c_{2},C) &=  \mu.Length (C)\; + \;\nu.Area (int(C)) \\
&\;+\; \lambda_{1}\int_{int(C)}\vert I(x,y)-c_{1}\vert^{2} dx dy \\
&+ \lambda_{2}\int_{ext(C)}\vert I(x,y)-c_{2}\vert^{2} dx dy 
\end{align*}
where $\mu\geq 0, \nu \geq 0 , \lambda_{1},\lambda_{2}> 0$ are fixed scalar parameters.
The level set evolution equation is given by
\begin{align*}
\frac{\partial \phi}{\partial t}= \delta_{\epsilon}(\phi)\left[\mu \;div\left(\frac{\nabla\phi}{\vert\nabla\phi\vert}\right) - \nu - \lambda_{1}\left(I-c_{1}\right)^{2} +\lambda_{2}\left(I-c_{2}\right)^{2}\right]
\end{align*}%
where $\delta_{\epsilon}$ is a smooth approximation of the Dirac delta function. A nice survey on active contours and level set implementation can be found in \cite{problems_in_image_processing}. We now describe our active contour model for texture segmentation.
\section{Proposed Active contour model for texture segmentation}
\label{sec:proposedalg}
In what follows, we assume familiarity with concepts from differential geometry like geodesic distance, Riemannian Exponential and Riemannian Logarithm maps. A thorough introduction to these concepts can be found in the books \cite{boothby1975,docarmo1992}.
We are given an intensity image $I:\Omega\subset \Rr^2 \rightarrow \Rr$. Our algorithm assumes that the image contains a background and a foreground texture. At every point $x\in \Omega$, let $N(x)$ be a $R^2 \times 1$ vector of intensities over a small neighborhood\footnote{Although we use a continuous region $\Omega$ to model the image domain, we are implicitly assuming a discrete image domain while defining the concept of a neighborhood $N(x)$. We choose to ignore this discrepancy.}, say of size $R \times R$. Given a closed contour $C$ on $\Omega$, we define the following two covariance matrices:
\begin{align}
\nonumber M^i(C) = & \frac{\int_{int(C)} N(x)N(x)^T\ dx}{\int_{int(C)} \ dx}\\
M^e(C) = & \frac{\int_{ext(C)} N(x)N(x)^T\ dx}{{\int_{ext(C)} \ dx}}
\label{eqn:mime}
\end{align}
where $int(C), ext(C)$ denote the interior and exterior of $C$ respectively. Note that $M^i(C)$ and $M^e(C)$ both belong to the set of $R^2 \times R^2$ symmetric positive definite matrices, which is a Riemannian manifold henceforth denoted by $PD(R^2)$. Let $d :PD(R^2) \times PD(R^2) \rightarrow \Rr$ denote the geodesic distance between two points of this manifold. Since the image contains two different texture regions, it is evident that the two covariance matrices(points on this manifold) defined in \eqref{eqn:mime} will be furthest away (in terms of geodesic distance) from each other when the contour $C$ lies on the boundary between the two textures. We justify this claim with an empirical evidence in Figure \ref{img:contourdistance}.
\begin{figure}[h]
    \centering
    \subfigure[]
    {
        \includegraphics[width=1.1 in]{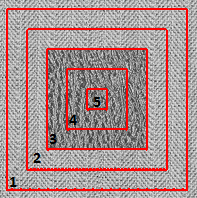}
    }
    \subfigure[]
    {
        \includegraphics[width=1.3 in]{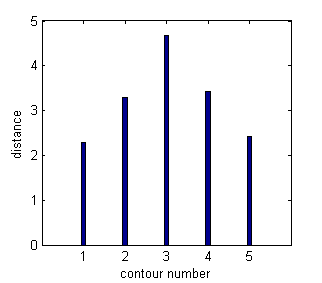}
    }
    \\
    \subfigure[]
    {
        \includegraphics[width=1.1 in]{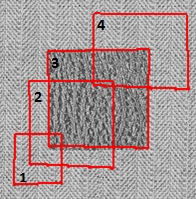}
    }
    \subfigure[]
    {
        \includegraphics[width=1.3 in]{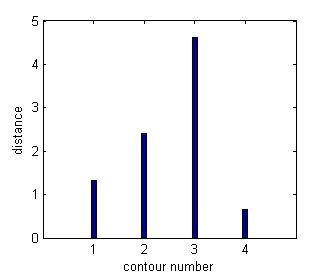}
    }
    \caption{(left) Different contours on an image with a foreground/background textures, (right) the corresponding (referred by appropriate contour number) geodesic distance between the covariance matrices defined in \eqref{eqn:mime}.} \label{img:contourdistance}
\end{figure}
For a given texture image $I:\Omega\rightarrow \Rr$, we propose the following cost function on the set of all closed contours defined on $\Omega$:
\begin{equation}
J(C) =  d(M^i(C),M^e(C))
\label{eqn:geocost}
\end{equation}
where $M^i(C),M^e(C)$ are defined in Equation \eqref{eqn:mime}. We find the contour $C$ that maximizes this cost, using the gradient ascent approach giving us a novel active contour scheme. Instead of working on parametric representations of $C$, we work with its level set representation which has several benefits as discussed in \cite{problems_in_image_processing}.\\
The curve $C$ is represented as the zero level set of the function $\phi:\Omega\rightarrow\Rr$, i.e., $C = \phi^{-1}(0)$. A typical choice of $\phi$ is the signed distance function of $C$:
\begin{align*}
\phi(x) = \left\{
\begin{array}{c l}     
    -d_C(x,C) & x \in int(C)\\
    d_C(x,C) & x \in \Omega \setminus int(C)
\end{array}\right.
\end{align*}
where 
\begin{equation*}
d_C(x,C) = \inf_{y \in C} d_{\Rr^2}(x,y)
\end{equation*}
with $d_{\Rr^2}(x,y)$ as the usual Euclidean distance on $\Rr^2$ between $x$ and $y$.
The sets $int(C),ext(C)$ can then be defined in terms of the level set function $\phi$ as 
\begin{align}
\nonumber int(C) = & \{x \in \Omega | \phi(x) < 0 \} \\
ext(C) = & \{x \in \Omega | \phi(x) \geq 0 \}.
\end{align}
Using the level set function $\phi$ and the Heaviside function
\begin{align*}
H(\phi) = \left\{
\begin{array}{c l}     
    1, &\ \phi \geq 0\\
    0, &\ \mbox{ otherwise},
\end{array}\right.
\end{align*}
we redefine the covariance matrices from Equation \eqref{eqn:mime}, as
\begin{align}
\nonumber M^{i}(\phi)= \frac{\int_{\Omega} (1-H(\phi)) N(x)N(x)^{T} dx}{\int_{\Omega}(1-H(\phi)) dx}\\
M^{e}(\phi)=\frac{\int_{\Omega} H(\phi) N(x)N(x)^{T} dx}{\int_{\Omega}H(\phi) dx}
\label{eqn:covlevelset}
\end{align}
Re-writing our cost function from Equation \eqref{eqn:geocost} in terms of the level set function $\phi$ gives us
\begin{equation}
 J(\phi)=d(M^{i}(\phi),M^{e}(\phi))
 \label{eqn:phicost}
 \end{equation}
To maximize this cost function we use gradient ascent algorithm, and the gradient is computed as follows
\begin{align}
\frac{\partial J}{\partial \phi}(\phi) =   \left\langle \frac{\partial d}{\partial M^{i}} , \frac{\partial M^{i}(\phi)}{\partial \phi}\right\rangle_{M^i}
+ \left\langle \frac{\partial d}{\partial M^e} , \frac{\partial M^e(\phi)}{\partial \phi}\right\rangle_{M^e}
\label{eqn:gradcost1}
\end{align}
where $\left\langle .,.\right\rangle_{M^i}$ and $\left\langle .,.\right\rangle_{M^e}$ are the Riemannian inner products defined on the Tangent space of $PD(R^2)$ at points $M^i(\phi)$ and $M^e(\phi)$, respectively. Specific details on this inner product can be found in \cite{fletcher2007}.
The derivatives of the geodesic distance $d$ is given by\footnote{A simpler explanation for this can be given in case we are working with $\Rr^2$ instead of $PD(R^2)$. In this case $\frac{\partial d}{\partial x}(x,y) = -(y-x)$ and $\frac{\partial d}{\partial y}(x,y) = -(x-y)$. This is exactly what is done by the Riemannian Log map on manifolds.}
\begin{align}
\label{eqn:graddist1}\frac{\partial }{\partial M^{i}} d (M^{i},M^{e})= -Log_{M^i}(M^e) \in T_{M^i}PD(R^2) \\
\frac{\partial }{\partial M^{e}} d (M^{i},M^{e})= -Log_{M^e}(M^i) \in T_{M^e}PD(R^2)
\label{eqn:graddist2}
\end{align}
where Log denotes the Riemannian log map defined on $PD(R^2)$. Derivatives of the covariance matrices defined in Equation \eqref{eqn:covlevelset} are given by
\begin{align}
\label{eqn:dmidp}\frac{\partial M^i}{\partial \phi}(\phi) & = \frac{1}{|\Omega_{int}|} \int_{\Omega} \left( M^i(\phi) -N(x)N(x)^{T}\right)\delta(\phi) dx \\
\frac{\partial M^e}{\partial \phi}(\phi) & = \frac{1}{|\Omega_{ext}|} \int_{\Omega} \left( N(x)N(x)^{T} - M^e(\phi) \right)\delta(\phi) dx 
\label{eqn:dmedp}
\end{align}
where $|\Omega_{int}|$ and $|\Omega_{ext}|$ are given by
\begin{align*}
|\Omega_{int}| & = \int_{\Omega} (1-H(\phi))dx \\
|\Omega_{ext}| & = \int_{\Omega} H(\phi)dx.
\end{align*}
and $\delta$ is the Dirac delta function.
Substituting Equations \eqref{eqn:graddist1},\eqref{eqn:graddist2},\eqref{eqn:dmidp},\eqref{eqn:dmedp} into Equation \eqref{eqn:gradcost1}, we get
\begin{align}
\nonumber\frac{\partial J}{\partial \phi} & = \int_{\Omega} \left[\left\langle -Log_{M^i}(M^e), \frac{1}{|\Omega_{int}|} \left( M^i(\phi) -N(x)N(x)^{T}\right)\delta(\phi) \right\rangle_{M^i} \right.\\ 
&\left. + \left\langle -Log_{M^e}(M^i),\frac{1}{|\Omega_{ext}|} \left( N(x)N(x)^{T} - M^e(\phi) \right)\delta(\phi) \right\rangle_{M^e}\right]\ dx
\end{align}
The gradient ascent as a level set evolution equation is therefore given by
\begin{align}
\nonumber\frac{\partial \phi}{\partial t}(x) & = \frac{\partial J}{\partial \phi}(x)\\
\nonumber& =\left\langle -Log_{M^i}(M^e), \frac{1}{|\Omega_{int}|} \left( M^i(\phi) -N(x)N(x)^{T}\right)\delta(\phi) \right\rangle_{M^i}\\ 
&+ \left\langle -Log_{M^e}(M^i),\frac{1}{|\Omega_{ext}|} \left( N(x)N(x)^{T} - M^e(\phi) \right)\delta(\phi) \right\rangle_{M^e}
\label{eqn:gradascent}
\end{align}
In the next section, we provide necessary implementation details and results on various images.
\section{Experiments}
\label{sec:expts}
The Heaviside and the Dirac delta functions are not continuous, instead, we use the following smooth approximation as given in \cite{chan2001}:
\begin{align}
H_{\epsilon}(\phi) = & \frac{1}{2}\left(1 + \frac{2}{\pi}\arctan\frac{\phi}{\epsilon}\right)\\
\delta_{\epsilon}(\phi) = & \frac{d}{d\phi} H_{\epsilon}(\phi)
\end{align}
We add curvature flow as a regularizer to
obtain the following level set evolution equation:
\begin{align}
\nonumber & \frac{\partial \phi}{\partial t}(x)  = \\
\nonumber& \left\langle -Log_{M^i}(M^e), \frac{1}{|\Omega_{int}|} \left( M^i(\phi) -N(x)N(x)^{T}\right)\delta(\phi) \right\rangle_{M^i}\\ 
\nonumber&+ \left\langle -Log_{M^e}(M^i),\frac{1}{|\Omega_{ext}|} \left( N(x)N(x)^{T} - M^e(\phi) \right)\delta(\phi) \right\rangle_{M^e} \\
& + \lambda \cdot div\left(\frac{\nabla \phi}{|\nabla \phi|}\right) \delta(\phi),
\label{eqn:final}
\end{align}
where $\kappa = div\frac{\nabla \phi}{|\nabla \phi|}$ is the curvature of the curve and $\lambda$ is the weight assigned to the curvature term.\\
We evolve an initial contour given by the user till the cost function given in Equation \eqref{eqn:geocost} increases, we stop the evolution the moment it decreases. We re-initialize the level set function when required following the algorithm given in \cite{sussman1994}. All images shown in this section are of size $200 \times 200$ pixels. The results shown in this section were obtained on a Intel Core2Duo, 2GB RAM machine using MATLAB. The time required for computing these results was under $10$ minutes, most of the time taken in re-initializing the level set function.\\
We begin by first validating our algorithm on artificially created images. Results are shown in Figure \ref{fig:arttextures}. 
\begin{figure}[htbp]
    \centering
    \subfigure[]
    {
        \includegraphics[width=1 in]{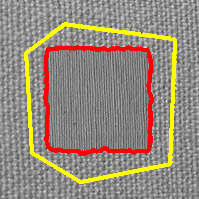}
        \label{arttexure1}
    }
    \subfigure[]
    {
        \includegraphics[width=1 in]{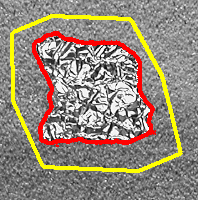}
        \label{arttexure2}
    }
    \subfigure[]
    {
        \includegraphics[width=1 in]{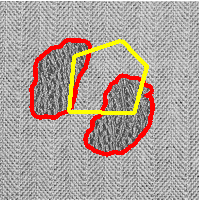}
        \label{arttexture3}
    }
    \caption{Segmentation results on artificial texture images. As is evident, the change of topology property is preserved by our model. The size of neighborhood for these results is $R = 5$ ($5 \times 5$ pixels)  here, i.e. the manifold under consideration is $PD(25)$. Initial contour in yellow, final contour in red.} 
\label{fig:arttextures}
\end{figure}
Topology of the evolving contour can change due to the level set implementation. Next, we give results on real texture images, that of a zebra and the Europe night sky image, in Figure \ref{fig:realtextures}. Texture being a neighborhood property rather than a pixel property, the segmented boundary will lie a pixel or two away from the actual boundary.
\begin{figure}[htbp]
    \centering
    \subfigure[]
    {
        \includegraphics[width=1.3 in]{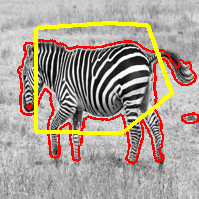}
        \label{realtexure1}
    }
    \subfigure[]
    {
        \includegraphics[width=1.3 in]{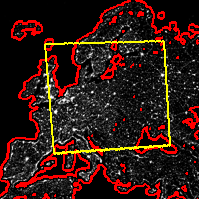}
        \label{realtexure2}
    }
    \caption{Segmentation results on real texture images. The size of neighborhood for these results is $R = 5$ ($5 \times 5$ pixels)  here, i.e. the manifold under consideration is $PD(25)$. Initial contour in yellow, final contour is shown in red.} 
\label{fig:realtextures}
\end{figure}

We next compare our results with the results generated by the algorithm in \cite{nawal_acm}, on some images from the Berkeley Segmentation dataset \cite{martin2001}, in Figure \ref{fig:comparewithKL}. We have used images from \cite{nawal_acm} to display their results. One can clearly see that our algorithm gives a better texture segmentation. Small noise-like artifacts are in fact regions where texture similar to the object texture is present, for instance, in the tiger image, there are reflection of the tiger strips in the water that our algorithm is able to successfully segment. 
\begin{figure}[htbp]
    \centering
    \subfigure[]
    {
        \includegraphics[width=1.3 in]{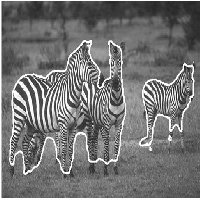}
        \label{kl1}
    }
    \subfigure[]
    {
        \includegraphics[width=1.3 in]{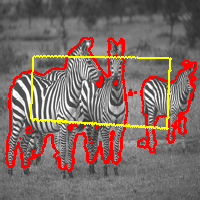}
        \label{our1}
    }\\
    \subfigure[]
    {
        \includegraphics[width=1.3 in]{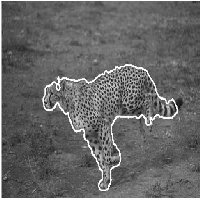}
        \label{kl2}
    }
    \subfigure[]
    {
        \includegraphics[width=1.3 in]{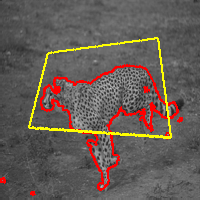}
        \label{our2}
    }\\
    \subfigure[]
    {
        \includegraphics[width=1.3 in]{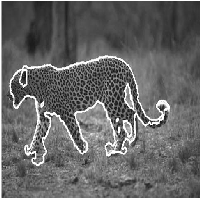}
        \label{kl3}
    }
    \subfigure[]
    {
        \includegraphics[width=1.3 in]{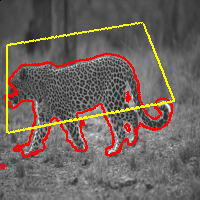}
        \label{our3}
    }\\
    \subfigure[]
    {
        \includegraphics[width=1.3 in]{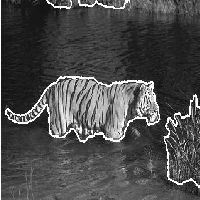}
        \label{kl4}
    }
    \subfigure[]
    {
        \includegraphics[width=1.3 in]{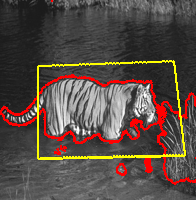}
        \label{our4}
    }
    \caption{Comparing results from \cite{nawal_acm}(left column) with results from our algorithm (right column). Initialized contour is shown in yellow, while the final contour in shown in red. The size of neighborhood for these results is $R = 5$ ($5 \times 5$ pixels)  here, i.e. the manifold under consideration is $PD(25)$. It can be seen that our results are better in most cases. The small noise-like artifacts are points around which object-like texture is present. For example, in the last image, our algorithm also captures the tiger-strips that appear due to reflection in water.}  
\label{fig:comparewithKL}
\end{figure}

With our algorithm, one can also segment usual gray level images, as explained next. Let the neighborhood size $R$ to be $1$, i.e. $N(x) = I(x)$. The covariance matrices $M^i(C), M^e(C)$ defined in Equation \eqref{eqn:mime}, will simply be the mean of squared intensities in the interior and exterior of the closed contour $C$, respectively. Also note that the covariance matrices now belong to $PD(1)$, i.e., the set of positive real numbers $\Rr^+$, of course with a metric different from the usual Euclidean one on $\Rr$. Our algorithm will then find the contour that maximizes the difference (geodesic distance on $PD(1)$) between the two numbers $M^i(C)$ and $M^e(C)$. Typical image segmentation results using this approach and results using the Chan \& Vese active contours \cite{chan2001} is shown in Figure \ref{fig:imgsegment}. 
\begin{figure}[htbp]
    \centering
    \subfigure[]
    {
        \includegraphics[width=1.3 in]{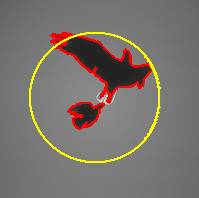}
    }
    \subfigure[]
    {
        \includegraphics[width=1.3 in]{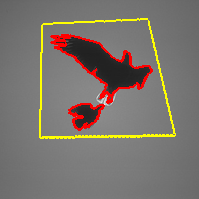}
    }
    \caption{Segmentation results on images using Chan \& Vese active contours (left) and our approach(right). The neighborhood size is $R=1$, as explained in the text. Initial contour in yellow, final contour in red.} 
\label{fig:imgsegment}
\end{figure}
Of course, with $R=5$, the covariance matrix can capture textures of that scale only. If we have large scale textures, our algorithm will over-segment the image, as can be seen in Figure \ref{fig:largescale}. Simply increasing the neighborhood size $R$ may not solve the problem, as the detected boundary may not be properly localized near the actual texture boundary.
\begin{figure}[htbp]
\centering
\includegraphics[width = 1.3in]{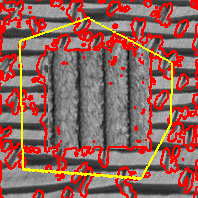}
\caption{The size of neighborhood for these results is $R = 5$ ($5 \times 5$ pixels)  here, i.e. the manifold under consideration is $PD(25)$. Our algorithm oversegments a texture image when both (foreground and background) textures under consideration are of larger scale than the neighborhood size used by our algorithm.}
\label{fig:largescale}
\end{figure}
\section{Conclusion}
In this paper, we propose a novel active contour based unsupervised texture segmentation algorithm. The algorithm finds a contour with maximum geodesic distance between its interior and exterior intensity covariance matrices. The results from previous section are in favor of our algorithm. With the least possible neighborhood size $R=1$, the process successfully segments gray-level images.\\
In its current state, the method depends on the size of the neighborhood $R$. Efforts are on to make it independent of $R$, either using a semi-supervised approach or using other multi-scale methods. Instead of intensity covariance matrices, one may also use covariance matrices of well-chosen multi-scale texture features. The method is able to capture even a slight deviation in a texture. This may be an advantage in some cases, but generally one may want the algorithm to be more invariant to small deviations.
\bibliographystyle{plain}
\bibliography{IJCV2013}
\end{document}